\documentclass{article}
\usepackage[accepted]{icml2018}
\usepackage{times}
\usepackage{courier}
\usepackage{graphicx}
\usepackage[utf8]{inputenc} % allow utf-8 input
\usepackage[T1]{fontenc}    % use 8-bit T1 fonts
\usepackage{hyperref}       % hyperlinks
\usepackage{url}            % simple URL typesetting
\usepackage{nicefrac}       % compact symbols for 1/2, etc.
\usepackage{amssymb}
\usepackage{amsmath}
\usepackage{amsthm}
\usepackage{paralist}
\usepackage{subfigure}
\usepackage{setspace}
\usepackage{booktabs}
\usepackage{microtype}
\usepackage{balance}

\usepackage{amssymb}
\usepackage{amsfonts}
\usepackage{mathrsfs}
\usepackage{xspace}
\usepackage{bm}
\usepackage{upgreek}

\newcommand{\safemath}[2]{\newcommand{#1}{\ensuremath{#2}\xspace}}

\safemath{\bma}{\mathbf{a}}
\safemath{\bmb}{\mathbf{b}}
\safemath{\bmc}{\mathbf{c}}
\safemath{\bmd}{\mathbf{d}}
\safemath{\bme}{\mathbf{e}}
\safemath{\bmf}{\mathbf{f}}
\safemath{\bmg}{\mathbf{g}}
\safemath{\bmh}{\mathbf{h}}
\safemath{\bmi}{\mathbf{i}}
\safemath{\bmj}{\mathbf{j}}
\safemath{\bmk}{\mathbf{k}}
\safemath{\bml}{\mathbf{l}}
\safemath{\bmm}{\mathbf{m}}
\safemath{\bmn}{\mathbf{n}}
\safemath{\bmo}{\mathbf{o}}
\safemath{\bmp}{\mathbf{p}}
\safemath{\bmq}{\mathbf{q}}
\safemath{\bmr}{\mathbf{r}}
\safemath{\bms}{\mathbf{s}}
\safemath{\bmt}{\mathbf{t}}
\safemath{\bmu}{\mathbf{u}}
\safemath{\bmv}{\mathbf{v}}
\safemath{\bmw}{\mathbf{w}}
\safemath{\bmx}{\mathbf{x}}
\safemath{\bmy}{\mathbf{y}}
\safemath{\bmz}{\mathbf{z}}
\safemath{\bmzero}{\mathbf{0}}
\safemath{\bmone}{\mathbf{1}}

\bmdefine{\biad}{a}
\bmdefine{\bibd}{b}
\bmdefine{\bicd}{c}
\bmdefine{\bidd}{d}
\bmdefine{\bied}{e}
\bmdefine{\bifd}{f}
\bmdefine{\bigd}{g}
\bmdefine{\bihd}{h}
\bmdefine{\biid}{i}
\bmdefine{\bijd}{j}
\bmdefine{\bikd}{k}
\bmdefine{\bild}{l}
\bmdefine{\bimd}{m}
\bmdefine{\bind}{n}
\bmdefine{\biod}{o}
\bmdefine{\bipd}{p}
\bmdefine{\biqd}{q}
\bmdefine{\bird}{r}
\bmdefine{\bisd}{s}
\bmdefine{\bitd}{t}
\bmdefine{\biud}{u}
\bmdefine{\bivd}{v}
\bmdefine{\biwd}{w}
\bmdefine{\bixd}{x}
\bmdefine{\biyd}{y}
\bmdefine{\bizd}{z}

\bmdefine{\bixid}{\xi}
\bmdefine{\bilambdad}{\lambda}
\bmdefine{\bimud}{\mu}
\bmdefine{\bithetad}{\theta}
\bmdefine{\biphid}{\phi}
\bmdefine{\bideltad}{\delta}

\safemath{\bmia}{\biad}
\safemath{\bmib}{\bibd}
\safemath{\bmic}{\bicd}
\safemath{\bmid}{\bidd}
\safemath{\bmie}{\bied}
\safemath{\bmif}{\bifd}
\safemath{\bmig}{\bigd}
\safemath{\bmih}{\bihd}
\safemath{\bmii}{\biid}
\safemath{\bmij}{\bijd}
\safemath{\bmik}{\bikd}
\safemath{\bmil}{\bild}
\safemath{\bmim}{\bimd}
\safemath{\bmin}{\bind}
\safemath{\bmio}{\biod}
\safemath{\bmip}{\bipd}
\safemath{\bmiq}{\biqd}
\safemath{\bmir}{\bird}
\safemath{\bmis}{\bisd}
\safemath{\bmit}{\bitd}
\safemath{\bmiu}{\biud}
\safemath{\bmiv}{\bivd}
\safemath{\bmiw}{\biwd}
\safemath{\bmix}{\bixd}
\safemath{\bmiy}{\biyd}
\safemath{\bmiz}{\bizd}

\safemath{\bmxi}{\bixid}
\safemath{\bmlambda}{\bilambdad}
\safemath{\bmmu}{\bimud}
\safemath{\bmtheta}{\bithetad}
\safemath{\bmphi}{\biphid}
\safemath{\bmdelta}{\bideltad}

\safemath{\bA}{\mathbf{A}}
\safemath{\bB}{\mathbf{B}}
\safemath{\bC}{\mathbf{C}}
\safemath{\bD}{\mathbf{D}}
\safemath{\bE}{\mathbf{E}}
\safemath{\bF}{\mathbf{F}}
\safemath{\bG}{\mathbf{G}}
\safemath{\bH}{\mathbf{H}}
\safemath{\bI}{\mathbf{I}}
\safemath{\bJ}{\mathbf{J}}
\safemath{\bK}{\mathbf{K}}
\safemath{\bL}{\mathbf{L}}
\safemath{\bM}{\mathbf{M}}
\safemath{\bN}{\mathbf{N}}
\safemath{\bO}{\mathbf{O}}
\safemath{\bP}{\mathbf{P}}
\safemath{\bQ}{\mathbf{Q}}
\safemath{\bR}{\mathbf{R}}
\safemath{\bS}{\mathbf{S}}
\safemath{\bT}{\mathbf{T}}
\safemath{\bU}{\mathbf{U}}
\safemath{\bV}{\mathbf{V}}
\safemath{\bW}{\mathbf{W}}
\safemath{\bX}{\mathbf{X}}
\safemath{\bY}{\mathbf{Y}}
\safemath{\bZ}{\mathbf{Z}}

\safemath{\bZero}{\mathbf{0}}
\safemath{\bOne}{\mathbf{1}}
\safemath{\bDelta}{\mathbf{\Delta}}
\safemath{\bLambda}{\mathbf{\Lambda}}
\safemath{\bPhi}{\mathbf{\Upphi}}
\safemath{\bSigma}{\mathbf{\Upsigma}}
\safemath{\bOmega}{\mathbf{\Upomega}}
\safemath{\bTheta}{\mathbf{\Uptheta}}

\bmdefine{\biAd}{A}
\bmdefine{\biBd}{B}
\bmdefine{\biCd}{C}
\bmdefine{\biDd}{D}
\bmdefine{\biEd}{E}
\bmdefine{\biFd}{F}
\bmdefine{\biGd}{G}
\bmdefine{\biHd}{H}
\bmdefine{\biId}{I}
\bmdefine{\biJd}{J}
\bmdefine{\biKd}{K}
\bmdefine{\biLd}{L}
\bmdefine{\biMd}{M}
\bmdefine{\biOd}{N}
\bmdefine{\biPd}{O}
\bmdefine{\biQd}{P}
\bmdefine{\biRd}{R}
\bmdefine{\biSd}{S}
\bmdefine{\biTd}{T}
\bmdefine{\biUd}{U}
\bmdefine{\biVd}{V}
\bmdefine{\biWd}{W}
\bmdefine{\biXd}{X}
\bmdefine{\biYd}{Y}
\bmdefine{\biZd}{Z}

\bmdefine{\biDelta}{\Delta}
\bmdefine{\biLambda}{\Lambda}
\bmdefine{\biPhi}{\Phi}
\bmdefine{\biSigma}{\Sigma}
\bmdefine{\biOmega}{\Omega}
\bmdefine{\biTheta}{\Theta}

\safemath{\bimA}{\biAd}
\safemath{\bimB}{\biBd}
\safemath{\bimC}{\biCd}
\safemath{\bimD}{\biDd}
\safemath{\bimE}{\biEd}
\safemath{\bimF}{\biFd}
\safemath{\bimG}{\biGd}
\safemath{\bimH}{\biHd}
\safemath{\bimI}{\biId}
\safemath{\bimJ}{\biJd}
\safemath{\bimK}{\biKd}
\safemath{\bimL}{\biLd}
\safemath{\bimM}{\biMd}
\safemath{\bimN}{\biNd}
\safemath{\bimO}{\biOd}
\safemath{\bimP}{\biPd}
\safemath{\bimQ}{\biQd}
\safemath{\bimR}{\biRd}
\safemath{\bimS}{\biSd}
\safemath{\bimT}{\biTd}
\safemath{\bimU}{\biUd}
\safemath{\bimV}{\biVd}
\safemath{\bimW}{\biWd}
\safemath{\bimX}{\biXd}
\safemath{\bimY}{\biYd}
\safemath{\bimZ}{\biZd}

\safemath{\bimDelta}{\biDelta}
\safemath{\bimLambda}{\biLambda}
\safemath{\bimPhi}{\biPhi}
\safemath{\bimSigma}{\biSigma}
\safemath{\bimOmega}{\biOmega}
\safemath{\bimTheta}{\biTheta}

\safemath{\setA}{\mathcal{A}}
\safemath{\setB}{\mathcal{B}}
\safemath{\setC}{\mathcal{C}}
\safemath{\setD}{\mathcal{D}}
\safemath{\setE}{\mathcal{E}}
\safemath{\setF}{\mathcal{F}}
\safemath{\setG}{\mathcal{G}}
\safemath{\setH}{\mathcal{H}}
\safemath{\setI}{\mathcal{I}}
\safemath{\setJ}{\mathcal{J}}
\safemath{\setK}{\mathcal{K}}
\safemath{\setL}{\mathcal{L}}
\safemath{\setM}{\mathcal{M}}
\safemath{\setN}{\mathcal{N}}
\safemath{\setO}{\mathcal{O}}
\safemath{\setP}{\mathcal{P}}
\safemath{\setQ}{\mathcal{Q}}
\safemath{\setR}{\mathcal{R}}
\safemath{\setS}{\mathcal{S}}
\safemath{\setT}{\mathcal{T}}
\safemath{\setU}{\mathcal{U}}
\safemath{\setV}{\mathcal{V}}
\safemath{\setW}{\mathcal{W}}
\safemath{\setX}{\mathcal{X}}
\safemath{\setY}{\mathcal{Y}}
\safemath{\setZ}{\mathcal{Z}}
\safemath{\emptySet}{\varnothing}

\safemath{\colA}{\mathscr{A}}
\safemath{\colB}{\mathscr{B}}
\safemath{\colC}{\mathscr{C}}
\safemath{\colD}{\mathscr{D}}
\safemath{\colE}{\mathscr{E}}
\safemath{\colF}{\mathscr{F}}
\safemath{\colG}{\mathscr{G}}
\safemath{\colH}{\mathscr{H}}
\safemath{\colI}{\mathscr{I}}
\safemath{\colJ}{\mathscr{J}}
\safemath{\colK}{\mathscr{K}}
\safemath{\colL}{\mathscr{L}}
\safemath{\colM}{\mathscr{M}}
\safemath{\colN}{\mathscr{N}}
\safemath{\colO}{\mathscr{O}}
\safemath{\colP}{\mathscr{P}}
\safemath{\colQ}{\mathscr{Q}}
\safemath{\colR}{\mathscr{R}}
\safemath{\colS}{\mathscr{S}}
\safemath{\colT}{\mathscr{T}}
\safemath{\colU}{\mathscr{U}}
\safemath{\colV}{\mathscr{V}}
\safemath{\colW}{\mathscr{W}}
\safemath{\colX}{\mathscr{X}}
\safemath{\colY}{\mathscr{Y}}
\safemath{\colZ}{\mathscr{Z}}

\safemath{\opA}{\mathbb{A}}
\safemath{\opB}{\mathbb{B}}
\safemath{\opC}{\mathbb{C}}
\safemath{\opD}{\mathbb{D}}
\safemath{\opE}{\mathbb{E}}
\safemath{\opF}{\mathbb{F}}
\safemath{\opG}{\mathbb{G}}
\safemath{\opH}{\mathbb{H}}
\safemath{\opI}{\mathbb{I}}
\safemath{\opJ}{\mathbb{J}}
\safemath{\opK}{\mathbb{K}}
\safemath{\opL}{\mathbb{L}}
\safemath{\opM}{\mathbb{M}}
\safemath{\opN}{\mathbb{N}}
\safemath{\opO}{\mathbb{O}}
\safemath{\opP}{\mathbb{P}}
\safemath{\opQ}{\mathbb{Q}}
\safemath{\opR}{\mathbb{R}}
\safemath{\opS}{\mathbb{S}}
\safemath{\opT}{\mathbb{T}}
\safemath{\opU}{\mathbb{U}}
\safemath{\opV}{\mathbb{V}}
\safemath{\opW}{\mathbb{W}}
\safemath{\opX}{\mathbb{X}}
\safemath{\opY}{\mathbb{Y}}
\safemath{\opZ}{\mathbb{Z}}
\safemath{\opZero}{\mathbb{O}}
\safemath{\identityop}{\opI}

\safemath{\veca}{\bma}
\safemath{\vecb}{\bmb}
\safemath{\vecc}{\bmc}
\safemath{\vecd}{\bmd}
\safemath{\vece}{\bme}
\safemath{\vecf}{\bmf}
\safemath{\vecg}{\bmg}
\safemath{\vech}{\bmh}
\safemath{\veci}{\bmi}
\safemath{\vecj}{\bmj}
\safemath{\veck}{\bmk}
\safemath{\vecl}{\bml}
\safemath{\vecm}{\bmm}
\safemath{\vecn}{\bmn}
\safemath{\veco}{\bmo}
\safemath{\vecp}{\bmp}
\safemath{\vecq}{\bmq}
\safemath{\vecr}{\bmr}
\safemath{\vecs}{\bms}
\safemath{\vect}{\bmt}
\safemath{\vecu}{\bmu}
\safemath{\vecv}{\bmv}
\safemath{\vecw}{\bmw}
\safemath{\vecx}{\bmx}
\safemath{\vecy}{\bmy}
\safemath{\vecz}{\bmz}

\safemath{\veczero}{\bmzero}
\safemath{\vecone}{\bmone}
\safemath{\vecxi}{\bmxi}
\safemath{\veclambda}{\bmlambda}
\safemath{\vecmu}{\bmmu}
\safemath{\vectheta}{\bmtheta}
\safemath{\vecphi}{\bmphi}
\safemath{\vecdelta}{\bmdelta}

\safemath{\matA}{\bA}
\safemath{\matB}{\bB}
\safemath{\matC}{\bC}
\safemath{\matD}{\bD}
\safemath{\matE}{\bE}
\safemath{\matF}{\bF}
\safemath{\matG}{\bG}
\safemath{\matH}{\bH}
\safemath{\matI}{\bI}
\safemath{\matJ}{\bJ}
\safemath{\matK}{\bK}
\safemath{\matL}{\bL}
\safemath{\matM}{\bM}
\safemath{\matN}{\bN}
\safemath{\matO}{\bO}
\safemath{\matP}{\bP}
\safemath{\matQ}{\bQ}
\safemath{\matR}{\bR}
\safemath{\matS}{\bS}
\safemath{\matT}{\bT}
\safemath{\matU}{\bU}
\safemath{\matV}{\bV}
\safemath{\matW}{\bW}
\safemath{\matX}{\bX}
\safemath{\matY}{\bY}
\safemath{\matZ}{\bZ}
\safemath{\matzero}{\bmzero}

\safemath{\matDelta}{\bDelta}
\safemath{\matLambda}{\bLambda}
\safemath{\matPhi}{\bPhi}
\safemath{\matSigma}{\bSigma}
\safemath{\matOmega}{\bOmega}
\safemath{\matTheta}{\bTheta}

\safemath{\matidentity}{\matI}
\safemath{\matone}{\matO}

\safemath{\rnda}{A}
\safemath{\rndb}{B}
\safemath{\rndc}{C}
\safemath{\rndd}{D}
\safemath{\rnde}{E}
\safemath{\rndf}{F}
\safemath{\rndg}{G}
\safemath{\rndh}{H}
\safemath{\rndi}{I}
\safemath{\rndj}{J}
\safemath{\rndk}{K}
\safemath{\rndl}{L}
\safemath{\rndm}{M}
\safemath{\rndn}{N}
\safemath{\rndo}{O}
\safemath{\rndp}{P}
\safemath{\rndq}{Q}
\safemath{\rndr}{R}
\safemath{\rnds}{S}
\safemath{\rndt}{T}
\safemath{\rndu}{U}
\safemath{\rndv}{V}
\safemath{\rndw}{W}
\safemath{\rndx}{X}
\safemath{\rndy}{Y}
\safemath{\rndz}{Z}

\safemath{\rveca}{\bimA}
\safemath{\rvecb}{\bimB}
\safemath{\rvecc}{\bimC}
\safemath{\rvecd}{\bimD}
\safemath{\rvece}{\bimE}
\safemath{\rvecf}{\bimF}
\safemath{\rvecg}{\bimG}
\safemath{\rvech}{\bimH}
\safemath{\rveci}{\bimI}
\safemath{\rvecj}{\bimJ}
\safemath{\rveck}{\bimK}
\safemath{\rvecl}{\bimL}
\safemath{\rvecm}{\bimM}
\safemath{\rvecn}{\bimN}
\safemath{\rveco}{\bomO}
\safemath{\rvecp}{\bimP}
\safemath{\rvecq}{\bimQ}
\safemath{\rvecr}{\bimR}
\safemath{\rvecs}{\bimS}
\safemath{\rvect}{\bimT}
\safemath{\rvecu}{\bimU}
\safemath{\rvecv}{\bimV}
\safemath{\rvecw}{\bimW}
\safemath{\rvecx}{\bimX}
\safemath{\rvecy}{\bimY}
\safemath{\rvecz}{\bimZ}

\safemath{\rvecxi}{\bmxi}
\safemath{\rveclambda}{\bmlambda}
\safemath{\rvecmu}{\bmmu}
\safemath{\rvectheta}{\bmtheta}
\safemath{\rvecphi}{\bmphi}

\safemath{\rmatA}{\bimA}
\safemath{\rmatB}{\bimB}
\safemath{\rmatC}{\bimC}
\safemath{\rmatD}{\bimD}
\safemath{\rmatE}{\bimE}
\safemath{\rmatF}{\bimF}
\safemath{\rmatG}{\bimG}
\safemath{\rmatH}{\bimH}
\safemath{\rmatI}{\bimI}
\safemath{\rmatJ}{\bimJ}
\safemath{\rmatK}{\bimK}
\safemath{\rmatL}{\bimL}
\safemath{\rmatM}{\bimM}
\safemath{\rmatN}{\bimN}
\safemath{\rmatO}{\bimO}
\safemath{\rmatP}{\bimP}
\safemath{\rmatQ}{\bimQ}
\safemath{\rmatR}{\bimR}
\safemath{\rmatS}{\bimS}
\safemath{\rmatT}{\bimT}
\safemath{\rmatU}{\bimU}
\safemath{\rmatV}{\bimV}
\safemath{\rmatW}{\bimW}
\safemath{\rmatX}{\bimX}
\safemath{\rmatY}{\bimY}
\safemath{\rmatZ}{\bimZ}

\safemath{\rmatDelta}{\bimDelta}
\safemath{\rmatLambda}{\bimLambda}
\safemath{\rmatPhi}{\bimPhi}
\safemath{\rmatSigma}{\bimSigma}
\safemath{\rmatOmega}{\bimOmega}
\safemath{\rmatTheta}{\bimTheta}

\usepackage{amssymb}
\usepackage{amsfonts}
\usepackage{mathrsfs}
\usepackage{xspace}
\usepackage{bm}
\usepackage{fancyref}
\usepackage{textcomp}

\usepackage{multirow}
\usepackage{stmaryrd}

\newenvironment{textbmatrix}{	\setlength{\arraycolsep}{2.5pt}%
								\big[\begin{matrix}}{\end{matrix}\big]%
								\raisebox{0.08ex}{\vphantom{M}}}

\def\be{\begin{equation}}
\def\ee{\end{equation}}
\def\een{\nonumber \end{equation}}
\def\mat{\begin{bmatrix}}
\def\emat{\end{bmatrix}}
\def\btm{\begin{textbmatrix}}
\def\etm{\end{textbmatrix}}

\def\ba#1\ea{\begin{align}#1\end{align}}
\def\bas#1\eas{\begin{align*}#1\end{align*}}
\def\bs#1\es{\begin{split}#1\end{split}} 
\def\bg#1\eg{\begin{gather}#1\end{gather}}
\def\bml#1\eml{\begin{multline}#1\end{multline}}
\def\bi#1\ei{\begin{itemize}#1\end{itemize}}

\newcommand{\lefto}{\mathopen{}\left}

\DeclareMathOperator{\tr}{tr}				% trace
\DeclareMathOperator{\sign}{sign}			% signum
\DeclareMathOperator{\kron}{\otimes}			% Kroneker Product
\DeclareMathOperator{\Exop}{\opE}			% expectation operator
\newcommand{\Ex}[2]{\ensuremath{\Exop_{#1}\lefto[#2\right]}} 	% expectation
\safemath{\dirac}{\delta}					% Dirac delta
\safemath{\krond}{\dirac}					% Kronecker delta
\safemath{\upto}{\uparrow}
\safemath{\downto}{\downarrow}
\safemath{\iu}{j}							% imaginary unit
\safemath{\ev}{\lambda}						% eigenvalue
\safemath{\hilseqspace}{l^{2}}				% Hilbert sequence space
\newcommand{\banachfunspace}[1]{\setL^{#1}}	% Banach function space
\safemath{\hilfunspace}{\banachfunspace{2}}	% Hilbert function space
\safemath{\SNR}{\textsf{SNR}} 				% signal to noise ratio
\safemath{\PAR}{\textsf{PAR}} 				% signal to noise ratio
\safemath{\No}{N_0}							% noise spectral density
\safemath{\Es}{E_s}							% energy per symbol
\safemath{\Eb}{E_b}							% energy per bit
\safemath{\EbNo}{\frac{\Eb}{\No}}
\safemath{\EsNo}{\frac{\Es}{\No}}

\DeclareMathOperator{\CHop}{\ensuremath{\opH}} % channel operator
\safemath{\tvir}{\rndh_{\CHop}}				% time-varying impulse response
\safemath{\tvtf}{\rndl_{\CHop}}				% 	-''- transfer function
\safemath{\spf}{\rnds_{\CHop}}				% spreading function
\safemath{\bff}{H_{\CHop}}					% bi-freuqency function

\safemath{\ircf}{r_{h}}						% impulse response correlation fn.
\safemath{\tftvcf}{r_{s}}					% scattering function
\safemath{\tfcf}{r_{l}}						% time-frequency correlation fn.
\safemath{\bfcf}{r_{H}}						% bi-frequency correlation fn.

\safemath{\tcorr}{c_h}						% time-correlation function
\safemath{\scf}{c_{s}}						% spreading function
\safemath{\tfcorr}{c_{l}}					% transfer-function correlation
\safemath{\fcorr}{c_{H}}						% frequency-correlation function

\safemath{\mi}{I}							% mutual information
\safemath{\capacity}{C}						% capacity

\safemath{\normal}{\mathcal{N}}			% normal distribution
\safemath{\jpg}{\mathcal{CN}}			% jointly proper Gaussian
\safemath{\mchain}{\leftrightarrow}		% Markov chain
\safemath{\dB}{\,\mathrm{dB}}
\safemath{\dBm}{\,\mathrm{dBm}}
\safemath{\Hz}{\,\mathrm{Hz}}
\safemath{\kHz}{\,\mathrm{kHz}}
\safemath{\MHz}{\,\mathrm{MHz}}
\safemath{\GHz}{\,\mathrm{GHz}}
\safemath{\s}{\,\mathrm{s}}
\safemath{\ms}{\,\mathrm{ms}}
\safemath{\mus}{\,\mathrm{\text{\textmu}s}}
\safemath{\ns}{\,\mathrm{ns}}
\safemath{\ps}{\,\mathrm{ps}}
\safemath{\meter}{\,\mathrm{m}}
\safemath{\mm}{\,\mathrm{mm}}
\safemath{\cm}{\,\mathrm{cm}}
\safemath{\m}{\,\mathrm{m}}
\safemath{\W}{\,\mathrm{W}}
\safemath{\mW}{\, \mathrm{mW}}
\safemath{\J}{\,\mathrm{J}}
\safemath{\K}{\,\mathrm{K}}
\safemath{\bit}{\,\mathrm{bit}}
\safemath{\nat}{\,\mathrm{nat}}

\safemath{\define}{\triangleq}			% definition

\safemath{\equivalent}{\sim}
\safemath{\distas}{\sim}					% distributed according to
\safemath{\sdiff}{\Delta}				% symmetric set difference

\safemath{\reals}{\mathbb{R}}
\safemath{\positivereals}{\reals_{+}}
\safemath{\integers}{\mathbb{Z}}
\safemath{\posint}{\integers_{+}}
\safemath{\naturals}{\mathbb{N}}
\safemath{\posnaturals}{\naturals_{+}}
\safemath{\complexset}{\mathbb{C}}
\safemath{\rationals}{\mathbb{Q}}

\newcommand*{\fancyrefapplabelprefix}{app}		% Appendix
\newcommand*{\fancyrefthmlabelprefix}{thm}		% Theorem
\newcommand*{\fancyreflemlabelprefix}{lem}		% Lemma
\newcommand*{\fancyrefcorlabelprefix}{cor}		% Corollary
\newcommand*{\fancyrefdeflabelprefix}{def}		% Definition
\newcommand*{\fancyrefproplabelprefix}{prop}		% Proposition
\newcommand*{\fancyrefobslabelprefix}{obs}		% Observation
\newcommand*{\fancyrefexmpllabelprefix}{exmpl}
\newcommand*{\fancyrefalglabelprefix}{alg}		% Algorithm

\frefformat{vario}{\fancyrefseclabelprefix}{Section~#1}
\frefformat{vario}{\fancyrefthmlabelprefix}{Theorem~#1}
\frefformat{vario}{\fancyreflemlabelprefix}{Lemma~#1}
\frefformat{vario}{\fancyrefcorlabelprefix}{Corollary~#1}
\frefformat{vario}{\fancyrefdeflabelprefix}{Definition~#1}
\frefformat{vario}{\fancyrefobslabelprefix}{Observation~#1}
\frefformat{vario}{\fancyreffiglabelprefix}{Fig.~#1}
\frefformat{vario}{\fancyrefapplabelprefix}{Appendix~#1} 
\frefformat{vario}{\fancyrefeqlabelprefix}{(#1)}
\frefformat{vario}{\fancyrefproplabelprefix}{Proposition~#1}
\frefformat{vario}{\fancyrefexmpllabelprefix}{Example~#1}
\frefformat{vario}{\fancyrefalglabelprefix}{Algorithm~#1}

\newtheorem{thm}{Theorem}
\newtheorem{cor}[thm]{Corollary}   % Turned off theorem numbering

\newtheorem{rem}{Remark}
\safemath{\dictab}{[\,\dicta\,\,\dictb\,]}

\safemath{\ysig}{\bmy}
\safemath{\ysighat}{\hat{\ysig}}
\safemath{\ysigdim}{M}
\safemath{\xsig}{\bmx}
\safemath{\xsigdim}{N}
\safemath{\nx}{n_x}
\safemath{\zsig}{\bmz}
\safemath{\zsigdim}{\ysigdim}
\safemath{\rsig}{\bmr}
\safemath{\Adict}{\bA}
\safemath{\Adicttilde}{\widetilde{\Adict}}
\safemath{\Adictdim}{\outputdim\times\xsigdim}
\safemath{\avec}{\bma}
\safemath{\avectilde}{\tilde{\avec}}
\safemath{\Bdict}{\bB}
\safemath{\Bdicttilde}{\widetilde{\Bdict}}
\safemath{\Cdict}{\bC}
\safemath{\cvec}{\bmc}
\safemath{\Ddict}{\bD}
\safemath{\Ddictdim}{\ysigdim\times\xsigdim}
\safemath{\dvec}{\bmd}
\safemath{\Ddicttilde}{\widetilde{\bD}}
\safemath{\Bonb}{\bB}
\safemath{\bvec}{\bmb}
\safemath{\Bonbdim}{\ysigdim\times\ysigdim}
\safemath{\noise}{\bmn}
\safemath{\noisedim}{\ysigim}
\safemath{\err}{\bme}
\safemath{\errdim}{\ysigdim}
\safemath{\errset}{\setE}
\safemath{\nerr}{n_e}
\safemath{\delop}{\bP_\errset}
\safemath{\delopc}{\bP_{{\errset}^c}}

\safemath{\cplxi}{\imath}
\safemath{\cplxj}{\jmath}
\safemath{\dict}{\matD}
\safemath{\inputdim}{N}		% number of columns of dictionary D
\safemath{\outputdim}{M}		%number of rows of dictionary D
\safemath{\sparsity}{S}	%sparsity
\safemath{\inputdimA}{{N_a}}	%total number of elements in dictionary A
\safemath{\inputdimB}{{N_b}}	%total number of elements in dictionary B
\safemath{\elemA}{{n_a}}	%number of elements chosen from dictionary A
\safemath{\elemB}{{n_b}}	%number of elements chosen from dictionary B
\safemath{\resA}{\matR_a}	%restriction map to elements of dictionary A
\safemath{\resB}{\matR_b}	%restriction map to elements of dictionary B
\safemath{\subD}{\matS} %subdictionary
\safemath{\subA}{\matS_a} %subdictionary part of A
\safemath{\subB}{\matS_b} %subdictionary part of B
\safemath{\dicta}{\matA} 	% first subdictionary
\safemath{\dictb}{\matB} 	% second subdictionary
\safemath{\hollowS}{H}
\safemath{\hollowA}{H_a}
\safemath{\hollowB}{H_b}
\safemath{\cross}{Z}
\safemath{\coh}{\mu_d}			% coherence dictionary
\safemath{\coha}{\mu_a}			% coherence first subdictionary
\safemath{\cohb}{\mu_b}			% coherence second subdictionary
\safemath{\mubs}{\nu}	%block sub-coherence
\safemath{\cohm}{\mu_m} %mutual coherence
\safemath{\dictset}{\setD}	% set of dictionaries
\safemath{\dictsetp}{\dictset(\coh,\coha,\cohb)}	% set of dictionaries parametrized
\safemath{\dictsetgen}{\dictset_\text{gen}}
\safemath{\dictsetgenp}{\dictsetgen(\coh)}
\safemath{\dictsetonb}{\dictset_\text{onb}}
\safemath{\dictsetonbp}{\dictsetonb(\coh)}

\safemath{\leftside}{U}
\safemath{\rightsideA}{R_a}
\safemath{\rightsideB}{R_b}

\safemath{\indexS}{\setI_S} %set of indices participating in sub-dictionary S

\safemath{\na}{n_a}			% cardinality of set of linearly independent columns of first dictionary
\safemath{\nb}{n_b}			% cardinality of set of linearly independent columns of second dictionary
\safemath{\coeffa}{p_i}	%coefficients for columns of A
\safemath{\coeffb}{q_j}	%coefficients for columns of B
\safemath{\seta}{\setP}		% set of linearly independent columns of A
\safemath{\setb}{\setQ}     % set of linearly independent columns of B
\safemath{\setw}{\setW}	%set of n largest elements of w
\safemath{\setz}{\setZ}	%set of L-n largest elements of z
\safemath{\cola}{\veca}		% generic element of the dictionary A
\safemath{\colb}{\vecb}		% generic element of the dictionary B
\safemath{\cold}{\vecd}		% generic element of the dictionary D
\safemath{\inputvec}{\vecx} 	%coefficient vector (input)
\safemath{\error}{\vece}	%error vector
\safemath{\noiseout}{\vecz} 	%noisy output vector
\safemath{\inputvecel}{x}
\safemath{\inputveca}{\vecx_a}
\safemath{\inputvecb}{\vecx_b}
\safemath{\outputvec}{\vecy}	%output of Dictionary
\safemath{\lambdamin}{\lambda_{\mathrm{min}}}
\safemath{\elltwo}{\ell_2}
\safemath{\ellone}{\ell_1}
\safemath{\ellzero}{\ell_0}
\safemath{\ellinf}{\ell_\infty}
\safemath{\ellinftilde}{\ell_{\widetilde\infty}}
\safemath{\licard}{Z(\coh,\coha,\cohb)}
\safemath{\xsol}{\hat{x}}
\safemath{\xbord}{x_b}		%Solution at the border
\safemath{\xstat}{x_s}		%Solution stationary in l0 prob
\safemath{\xstatLone}{\tilde{x}_s}
\safemath{\order}{\mathcal{O}} %order notation (big O)
\safemath{\scales}{\Theta} %scales as
\safemath{\ones}{\mathbf{1}} %all ones matrix
\safemath{\zeroes}{\mathbf{0}} %all zeroes matrix
\safemath{\thlone}{\kappa(\coh,\cohb)} %treshold l1 problem
\safemath{\constoneA}{\delta} %constant in l1 theorem to save space
\safemath{\constoneB}{\epsilon} %constant in l1 theorem to save space
\safemath{\nlarge}{L}				   %num large elements
\safemath{\sumlarge}{S_\nlarge}
\safemath{\maxlarger}{P_\nlarge}	   % maximum in Gribonval and Nielsen
\safemath{\Pzero}{\textrm{P0}}	
\safemath{\Pone}{\textrm{P1}}
\safemath{\vecfir}{\vecw}			 % \vecv element of the kernel of the dictionary, \vecv=[\vecfir \vecsec]
\safemath{\vecsec}{\vecz}
\safemath{\elvecfir}{w}              % element of vecfir
\safemath{\elvecsec}{z}				 % element of vecsec
\safemath{\nlargefir}{n}
\safemath{\normout}{\gamma}
\safemath{\auxfun}{h}
\safemath{\supp}{\textrm{supp}}%support

\safemath{\indexa}{\ell}
\safemath{\indexb}{r}
\safemath{\indexc}{i}
\safemath{\indexd}{j}

\safemath{\project}{P}%projector

\allowdisplaybreaks

\usepackage{natbib}
\usepackage{color}
\icmltitlerunning{An Estimation and Analysis Framework for the Rasch Model}

\begin{document}

\twocolumn[
\icmltitle{An Estimation and Analysis Framework for the Rasch Model}

\icmlsetsymbol{equal}{*}

\begin{icmlauthorlist}
	\icmlauthor{Andrew S. Lan}{pr}
	\icmlauthor{Mung Chiang}{pd}
	\icmlauthor{Christoph Studer}{cu}
\end{icmlauthorlist}

\icmlaffiliation{cu}{School~of Electrical and Computer Engineering, Cornell University}
\icmlaffiliation{pr}{Department of Electrical Engineering, Princeton University}
\icmlaffiliation{pd}{Purdue University}

\icmlcorrespondingauthor{Andrew S. Lan}{andrew.lan@princeton.edu}

\icmlkeywords{Linear estimation, Rasch model}

\vskip 0.3in
]



\printAffiliationsAndNotice{}

% ================================================================================
% ================================================================================
% ================================================================================
%%\vspace{-1.1cm}
\begin{abstract}
The Rasch model is widely used for item response analysis in applications ranging from  recommender systems to psychology, education, and finance.   
While a number of estimators have been proposed for the Rasch model over the last decades, the available analytical performance guarantees are mostly asymptotic. 
This paper provides a framework that relies on a novel linear minimum mean-squared error (L-MMSE) estimator which enables an exact, nonasymptotic, and closed-form  analysis of the parameter estimation error under the Rasch model.  
The proposed framework provides guidelines on the number of items and responses required to attain low estimation errors in tests or surveys. 
We furthermore demonstrate its efficacy on a number of real-world collaborative filtering datasets, which reveals that the proposed L-MMSE estimator performs on par with state-of-the-art nonlinear estimators in terms of predictive performance. 
\end{abstract}

% ================================================================================
% ================================================================================
% ================================================================================

%%%%%%%%%%%%%%%%%%%%%%%%%%%%%%%%%%%%%%%%%%%%%%%%%%%%%%%%%%

%\input{1-introduction.tex}
%\input{2-mainresults.tex}
% !TEX root = 18ICML_linrasch.tex
%%%%%%%%%%%%%%%%%%%%%%%%%%%%%%

\section{Introduction}

This paper presents a novel framework that enables an exact, nonasymptotic, and closed-form analysis of the parameter estimation error under the Rasch model. 
The Rasch model was proposed in 1960 for modeling the responses of students/users to test/survey items \cite{rasch}, and has enjoyed great success in applications including (but not limited to) psychometrics \cite{irtbook}, educational tests \cite{db}, crowdsourcing \cite{jake}, public health \cite{publichealth}, and even market and financial research \cite{finance,market}.  
Mathematically, the (dichotomous) Rasch model, also known as the 1PL item response theory (IRT) model \cite{lordirt}, is given by
\begin{align} \label{eq:raschmodel}
p(Y_{u,i} = 1) = \Phi(a_u - d_i),
\end{align} 
where $Y_{u,i} \in \{-1,+1\}$ denotes the response of user $u$ to item $i$, where $+1$ stands for a correct response and $-1$ stands for an incorrect response. The parameters $a_u \in \reals$ model the scalar \emph{abilities} of users~$u=1,\ldots,U$ and the parameters $d_i \in \reals$ model the scalar \emph{difficulties} of items~$i=1,\ldots,Q$.  The function $\Phi(x)=\int_{-\infty}^x \mathcal{N}(t;0,1)\text{d}t$, often referred to as the \emph{inverse probit link} function\footnote{While some publications assume the inverse logit link function, i.e., the sigmoid $\Phi(x)=\frac{1}{1 + e^{-x}}$, in most real-world applications the choice of the link function has no significant performance impact. In what follows, we will focus on the inverse probit link function for reasons that will be discussed in \fref{sec:mainresults}.}, is the cumulative distribution function of a standard normal random variable, where $\mathcal{N}(t;0,1)$ denotes the probability density function of a standard normal random variable evaluated at $t$.

The literature describes a range of parameter estimation methods under the Rasch model and related IRT models; see \cite{irtest} for an overview.  
However, existing analytical results for the associated \emph{parameter estimation error} are limited; see \cite{firsterr} for an example. 
The majority of existing results have been proposed in the psychometrics and educational measurement literature; see, e.g., \cite{errorbook} for a survey.
The proposed analysis tools rely, for example, on multiple imputation \cite{scaleerr} or Markov chain Monte Carlo (MCMC) techniques \cite{mcmcerr}, and are thus not analytical.  Hence, their accuracy strongly depends on the available data.

Other analysis tools use the Fisher information matrix \cite{3plerr,scaleerr} to obtain \emph{lower bounds} on the estimation error.  Such methods are of \emph{asymptotic} nature, i.e., they yield accurate results only when the number of users and items tend to infinity.  For real-world settings with limited data, these bounds are typically loose;  As an example, in computerized adaptive testing (CAT) \cite{seqtest}, a user enters the system and starts responding to items.  The system maintains an estimate of their ability parameter, and adaptively selects the next-best item to assign to the user that is most informative of the ability estimate.  Calculating the informativeness of each item requires an analysis of the uncertainty in the ability estimate.  Initially, after the user has only responded to a few items, these asymptotic methods lead to highly inaccurate analyses, which may lead to poor item selections. 

Another family of analysis tools relies on concentration inequalities and yield \emph{probabilistic} bounds, i.e., bounds that hold with high probability \cite{honestlogit,glmbandits}.  Such results are often impractical in real-world applications.  
However, an exact analysis of the estimation error of the Rasch model is critical to ensure the a certain degree of reliability of assessment scores in tests \cite{reliability}.

\subsection{Contributions}
We propose a novel framework for the Rasch model that enables an \emph{exact}, \emph{nonasymptotic}, and \emph{closed-form} analysis of the parameter estimation error.  
To this end, we generalize a recently-proposed linear estimator for binary regression~\cite{linprobitciss} to the Rasch model, which enables us to derive a \emph{sharp upper bound} on the mean squared error (MSE) of model parameter estimates. 
Our analytical results are in stark contrast to existing analytical results which either provide loose lower bounds or are asymptotic in nature, rendering them impractical in real-world applications.  

To demonstrate the efficacy of our framework, we provide experimental results on both synthetic and real-world data.
First, using synthetic data, we show that our upper bound on the MSE is (often significantly) tighter than the Fisher information-based lower bound, especially when the problem size is small and when the data is noisy.  Therefore, our framework enables a more accurate analysis of the estimation error in real-world settings.  
Second, using real-world student question response and user movie rating datasets, we show that our linear estimator achieves competitive predictive performance to more sophisticated, nonlinear estimators for which sharp performance guarantees are unavailable.

\section{Rasch Model and Probit Regression}
\label{sec:background}

The Rasch model in \fref{eq:raschmodel} can be written in equivalent matrix-vector form as follows \cite{hoffbook}:
\begin{align} \label{eq:probitmodel}
\bmy = \sign(\bD\bmx + \bmw).
\end{align}
Here, the $UQ$-dimensional vector $\bmy \in \{-1,+1\}^{U Q}$ contains all user responses to all items, the Rasch model matrix $\bD=[\bOne_{Q}\kron\bI_{U\times U},\bI_{Q\times Q}\kron\bOne_U ]$ is constructed with the Kronecker product operator $\kron$, identity matrices $\bI$, all-ones vectors $\bOne$, and the vector $\bmx^T=[\bma^T,-\bmd^T]$ to be estimated consists of the user abilities $\bma \in \mathbb{R}^U$ and item difficulties $\bmd \in \mathbb{R}^Q$. The ``noise'' vector $\bmw$ contains i.i.d.\ standard normal random variables.
In this equivalent form, parameter estimation under the Rasch model can be casted as a \emph{probit regression} problem \cite{probitml}, for which numerous estimators have been proposed in the past.

\subsection{Estimators for Probit Regression}

The two most prominent estimators for probit regression are the posterior mean (PM) estimator, given by 
\begin{align} \label{eq:PME}
\hat\bmx^\text{PM} =\textstyle  \Ex{\bmx}{\bmx| \bmy} = \int_{\reals^N} \bmx p(\bmx|\bmy)\text{d}\bmx,
\end{align}
and the maximum a-posteriori (MAP) estimator, given by
\begin{align*} %\label{eq:MAPestimator}
\hat\bmx^\text{MAP} = \textstyle \underset{\bmx\in\reals^N}{\text{arg\,min}}   -\!\sum_{m=1}^M \log(\Phi(y_m \bmd_m^T \bmx)) + \frac{1}{2} \bmx^T \bC_{\bmx}^{-1} \bmx.
\end{align*}
Here, $p(\bmx|\bmy)$ denotes the posterior probability of the vector~$\bmx$ given the observations $\bmy$ under the model  \fref{eq:probitmodel}, $\bmd^T_m$ denotes the $m$th row of the matrix of covariates $\bD$, and $\bC_\bmx$ denotes the covariance matrix of the multivariate Gaussian prior on $\bmx$.  
A special case of the MAP estimator is the well-known maximum likelihood (ML) estimator, which does not impose a prior distribution on $\bmx$. 

The PM estimator is optimal in terms of minimizing the MSE of the estimated parameters, which is defined as
\begin{align} \label{eq:MSE}
\textit{MSE}(\hat\bmx) = \Ex{\bmx,\bmw}{\|\bmx-\hat\bmx\|^2}.
\end{align} 
However, there are no simple methods to evaluate the expectation in~\fref{eq:PME} under the probit model. Thus, one typically resorts to Markov chain Monte Carlo (MCMC) methods \cite{albertchib} to perform PM estimation, which can be computationally intensive. 
In contrast to the PM estimator, MAP and ML estimation is generally less complex since it can be implemented using standard convex optimization algorithms \cite{wrightbook,tibsbook,goldstein2014field}. 
On the flipside, MAP and ML estimation is not optimal in terms of minimizing the MSE in \fref{eq:MSE}.
In contrast to such well-established, nonlinear estimators, we build our framework on the family of \emph{linear estimators} recently proposed in \cite{linprobitciss}.
There, a linear minimum MSE (L-MMSE) estimator was proposed for a certain class of probit regression problems.
This L-MMSE estimator was found to perform on par with the PM estimator and outperforms the MAP estimator in terms of the MSE for certain settings, while enabling an exact and nonasymptotic analysis of the MSE.

\subsection{Analytical Performance Guarantees}
In the statistical estimation literature, there exists numerous analytical results characterizing the estimation errors for binary regression problems in the \emph{asymptotic} setting.
For example, \cite{brillinger1982generalized} shows that least squares estimation is particularly effective when the design matrix~$\bD$ has i.i.d.\ Gaussian entries and the number of observations approaches infinity; in this case, its performance was shown to differ from that of the PM estimator only by a constant factor. 
Recently, \cite{thrampoulidis2015lasso} provides a related analysis in the case that the parameter vector $\bmx$ is sparse.  
Another family of \emph{probabilistic} results relies on the asymptotic normality property of ML estimators, either in the standard (dense) setting \cite{mlasymptoticlogit,mlasymptotoicglm} or the sparse setting \cite{honestlogit,bachlogit,wainwrightising,plan1bit}, providing bounds on the MSE with high probability. 
Since numerous real-world applications, such as the Rasch model, rely on deterministic, structured matrices and have small problem dimensions, existing analytical performance bounds are often loose; see \fref{sec:numericalresults} for experiments that support this claim.

\section{Main Results}
\label{sec:mainresults}

Our main result is as follows; the proof is given in \fref{app:pfknown}. 
\begin{thm} \label{thm:lmmseestimator}
Assume that $\bmx \distas \mathcal{N}(\bar{\bmx},\bC_\bmx)$ with mean vector~$\bar\bmx$ and positive definite covariance matrix $\bC_\bmx$, and assume that the vector $\bmw$ contains i.i.d.\ standard normal random variables.
Consider the general probit regression model 
\begin{align} \label{eq:probitmodelgeneral}
\bmy = \sign(\bD\bmx + \bmm + \bmw),
\end{align}
where $\bD$ is a given matrix of covariates and $\bmm$ is a given bias vector. Then, the L-MMSE estimate is given by
\begin{align*}
\hat\bmx^\text{\em L-MMSE} = \bE^T \bC_\bmy^{-1}\bmy + \bmb,
\end{align*}
where we use the following quantities:  
\begin{align*}
\bE & \! =\!  2 \mathrm{diag}(\mathcal{N}(\bmc;0,\!1) \! \odot \! \mathrm{diag}(\bC_\bmz)^{-\frac{1}{2}}) \bD \bC_\bmx  \\
\bmc & = \bar{\bmz} \odot \mathrm{diag}(\bC_\bmz)^{-1/2} \\
\bar{\bmz} & = \bD \bar{\bmx} + \bmm \\
\bC_\bmz & = \bD \bC_\bmx \bD^T + \bI \\
\bC_{\bmy} &  = 2(\Phi_2(\bmc\bOne^T, \bOne\bmc^T; \bR) + \Phi_2(-\bmc\bOne^T, -\bOne\bmc^T; \bR)) \\
& \quad - \bOne_{M\times M}  -  \bar{\bmy} \bar{\bmy}^T \\
\bR & = \mathrm{diag}( \mathrm{diag}(\bC_\bmz)^{-1/2}) \bC_\bmz \mathrm{diag}( \mathrm{diag}(\bC_\bmz)^{-1/2}) \\
\bar{\bmy} & \! = \! \Phi(\bmc) - \Phi(-\bmc) \\
\bmb & \!= \! \bar{\bmx} - \bE^T \bC_\bmy^{-1}\bar{\bmy}.
\end{align*}
Here, $\Phi_2(x, y, \rho)$ denotes the cumulative density of a two-dimensional zero-mean Gaussian distribution with covariance matrix $[1\; \rho; \rho \; 1]$ with $\rho\in[0,1)$, defined as
\begin{align*}
\Phi_2(x, y; \rho) = \int_{-\infty}^x \int_{-\infty}^y \frac{1}{2 \pi \sqrt{1 - \rho^2}} e^{-\frac{s^2 - 2\rho st + t^2}{2(1 - \rho^2)}} \mathrm{d}t \mathrm{d}s
\end{align*}
and is applied element-wise on matrices. Furthermore, the associated estimation MSE is given by 
\begin{align*}
\textit{MSE}(\hat\bmx^\text{\em L-MMSE}) = \tr(\bC_\bmx - \bE^T \bC_\bmy^{-1} \bE).
\end{align*}
\end{thm}

We note that the linear estimator developed in \citep[Thm.~1]{linprobitciss} is a special case of our result with $\bar\bmx = \bZero$ and $\bmm = \bZero$. As we will show below, including both of these terms will be essential for our analysis.

\begin{rem}
We exclusively focus on probit regression since the matrices $\bE$ and $\bC_{\bmy}$ exhibit tractable expressions under this model. We are unaware of any closed-form expressions for these quantities in the logistic regression case.
\end{rem}

As an immediate consequence of the fact that the PM estimator minimizes the MSE, we can use \fref{thm:lmmseestimator} to obtain the following upper bound on the MSE of the PM estimator.
\begin{cor} \label{cor:mseoflmmseestimator}
The MSE of the PM estimator is upper-bounded as follows:
\begin{align} \label{eq:PMbound}
\textit{MSE}(\bmx^\text{PM})   \leq  \textit{MSE}(\hat\bmx^\text{\em L-MMSE}).
\end{align}
\end{cor}

As we will demonstrate in \fref{sec:numericalresults}, this upper bound on the MSE turns out to be surprisingly sharp for a broad range of parameters and problem settings.

We now specialize \fref{thm:lmmseestimator} for the Rasch model and use  \fref{cor:mseoflmmseestimator} to analyze the associated MSE.  We divide our results into two cases: (i) both the user abilities and item difficulties are unknown and (ii) one of the two sets of parameters is known and the other is unknown.  Due to symmetry in the Rasch model, we will present our results with unknown/known item difficulties while the user abilities are unknown and to be estimated; a corresponding analysis on the estimation error of item parameters follows immediately.

\subsection{First Case: Unknown Item Parameters}
\label{sec:raschuk}

We now analyze the case in which both the user abilities and item difficulties are unknown and need to be estimated.  In practice, this scenario is relevant if a new set of items are deployed with little or no prior knowledge on their difficulty parameters.  
We assume that there is no missing data, i.e., we observe all user responses to all items.\footnote{Our analysis can readily be generalized to missing data; the results, however, depend on the missing data pattern.}
In the psychometrics literature (see, e.g.,~\cite{raschest}), one typically assumes that the entries of the ability~$\bma$ and difficulty vectors~$\bmd$ are i.i.d.\ zero-mean Gaussian with variance $\sigma^2_a$ and $\sigma^2_d$, respectively, i.e., $a_u\sim\setN(0,\sigma^2_a)$ and $d_i\sim\setN(0,\sigma^2_d)$, which can be included in our model assumptions.
Thus, we can leverage the special structure of the Rasch model, since it corresponds to a special case of the generic probit regression model in \fref{eq:probitmodelgeneral} with $\bD = [\bOne_{Q}\kron\bI_{U\times U},\bI_{Q\times Q}\kron\bOne_U]$ and $\bmm = \bZero$.  
We have the following result on the MSE of the L-MMSE estimator; the proof is given in \fref{app:pfunknown}. 
\begin{thm} \label{thm:rasch}
Assume that $\sigma^2_a = \sigma^2_d = \sigma^2_x$ and the covariance matrix of $\bmx$ is $\bC_\bmx = \sigma^2_x \bI_{(U+Q) \times (U+Q)}$. Let 
\begin{align*}
s= \frac{2}{\pi} \arcsin\!\left(\frac{\sigma^2_x}{2\sigma^2_x + 1}\right).
\end{align*} 
Then, the MSE of the L-MMSE estimator of user abilities under the Rasch model is given by
\begin{align} \label{eq:mseavg} 
\notag &  \textit{MSE}_a = \Ex{\bmx,\bmw}{(a_u - \hat{a}_u)^2} = \\
& \sigma_x^2 \! \left(\!1 \!-\! \frac{2}{\pi} \frac{\sigma^2_x}{2 \sigma^2_x + 1} \frac{sQ(Q+U-3)+1}{(s(Q-2)+1)(s(Q+U-2)+1)}\!\right)\!.
\end{align}
\end{thm}

To the best of our knowledge, \fref{thm:rasch} is the first exact, nonasymptotic, and closed-form analysis of the MSE of a parameter estimation method for the Rasch model.  
From~(\ref{eq:mseavg}), we see that if $\sigma^2_x$ is held constant, then the relationship between $\textit{MSE}_a$ and the numbers of users ($U$) and items ($Q$) is given by the ratio of two second-order polynomials.  
If the signal-to-noise ratio (SNR) is low (or, equivalently, the data is noisy), i.e., $\sigma^2_x \ll \sigma^2_n$, then we have $\frac{\sigma^2_x}{2\sigma^2_x + 1} \approx 0$ and hence, $s = \frac{2}{\pi} \arcsin(\frac{\sigma^2_x}{2\sigma^2_x + 1}) \approx 0$.  In this case, we have $\text{MSE}_a \approx \sigma^2_x$, i.e., increasing the number of users/items does not affect the accuracy of the ability and difficulty parameters of users and items; this behavior is as expected.

When $U,Q \rightarrow \infty$, the MSE satisfies  
\begin{align}
\text{MSE}_a \rightarrow  \sigma^2_x \!\left(1 - \frac{\sigma^2_x}{2\sigma^2_x + 1} \arcsin^{-1}\! \left(  \frac{\sigma^2_x}{2\sigma^2_x + 1}\right)\!\right)\!,
\end{align}
which is a non-negative quantity. 
This result implies that the L-MMSE estimator has a residual MSE even as the number of users/items grows large. More specifically, since  $x \leq \arcsin(x)$ for $x \in [0,1]$, this residual error approaches $\sigma^2_x (1 - \frac{3}{\pi})$ at high values of SNR.  We note, however, this result does not imply that the L-MMSE estimator is not consistent under the Rasch model, since the number of parameters to be estimated ($U + Q$) grows with the number of the observations~($UQ$) instead of remaining constant.

\begin{rem}
The above MSE analysis is \emph{data-independent}, in contrast to error estimates that rely on the responses $\bmy$ (which is, for example, the case for method in \cite{errorbook}).  This fact implies that our result provides an error estimate \emph{before} observing $\bmy$.  Thus, \fref{thm:rasch} provides guidelines on how many items to include and how many users to recruit for a study, given a desired MSE level on the user ability and item difficulty parameter estimates.
\end{rem}

%\newpage
\subsection{Second Case: Known Item Difficulties}
\label{sec:raschk}

We now analyze the case in which the user abilities are unknown and need to be estimated; the item difficulties ($\bmd$) are given.
In practice, this scenario is relevant if a large number of users previously responded to a set of items so that a good estimate of the item difficulties is available.  
Let~$a$ denote the scalar ability parameter of an user. Then, their responses to items are modeled as
\begin{align*} 
p(\bmy = 1) = \Phi(\bOne_{Q} a - \bmd).
\end{align*} 
The following result follows from \fref{thm:lmmseestimator} by setting $\bmx = a$, $\bar{\bmx} = \bar{x}$, $\bC_\bmx = \sigma^2_x$, $\bD = \bOne_Q$, and $\bmm = -\bmd$. 
\begin{cor} \label{thm:raschk}
Assume that $a \distas \mathcal{N}(\bar{x},\sigma^2_x)$. Then, the L-MMSE estimate of user ability is given by
\begin{align*}
\hat a = \bme^T \bC_{\bmy}^{-1} \bmy + b,
\end{align*}
where  
\begin{align*}
\bme & \!=\! 2 \frac{\sigma^2_x}{\sqrt{\sigma^2_x + 1}} \mathcal{N}(\bmc;0,1) \\
\bmc & = \bar{\bmz} \odot \mathrm{diag}(\bC_\bmz)^{-1/2} \\
\bar{\bmz} & = \bar{x} \bOne_Q - \bmd \\
\bC_\bmz & = \sigma^2_x \bOne_{Q \times Q} + \bI \\
\bar{\bmy} & = \Phi(\bmc) - \Phi(-\bmc) \\
\bC_{\bmy} & =  2(\Phi_2(\bmc\bOne^T\!\!, \bOne\bmc^T, \bR) + \Phi_2(-\bmc\bOne^T, -\bOne\bmc^T, \bR)) \\
& \quad - \bOne_{M \times M} - \bar{\bmy} \bar{\bmy}^T \\
\bR & = \mathrm{diag}( \mathrm{diag}(\bC_\bmz)^{-1/2}) \bC_\bmz \mathrm{diag}( \mathrm{diag}(\bC_\bmz)^{-1/2}.
\end{align*}
The MSE of the user ability estimate is given by $\textit{MSE}(\hat a) =\sigma_x^2 - \bme^T \bC_{\bmy}^{-1} \bme$. 
\end{cor}

% !TEX root = 18ICML_linrasch.tex
%%%%%%%%%%%%%%%%%%%%%%%%%%%%%%

\section{Numerical Results}
\label{sec:numericalresults}

We now experimentally demonstrate the efficacy of the proposed framework.
First, we use synthetically generated data to numerically compare our L-MMSE-based upper bound on the MSE of the PM estimator to the widely-used lower bound based on Fisher information \cite{3plerr,scaleerr}. 
We then use several real-world collaborative filtering datasets to show that the L-MMSE estimator achieves comparable predictive performance to that of the PM and MAP estimators.

\subsection{Experiments with Synthetic Data}

We start with synthetic data to demonstrate the exact and nonasymptotic nature of our analytical MSE expressions. 

\subsubsection{First Case: Unknown Item Parameters}

\begin{figure*}[t]
\vspace{-0.4cm}
\centering
\subfigure[$U=20$, SNR$=-10$dB.]{
\includegraphics[scale=0.28]{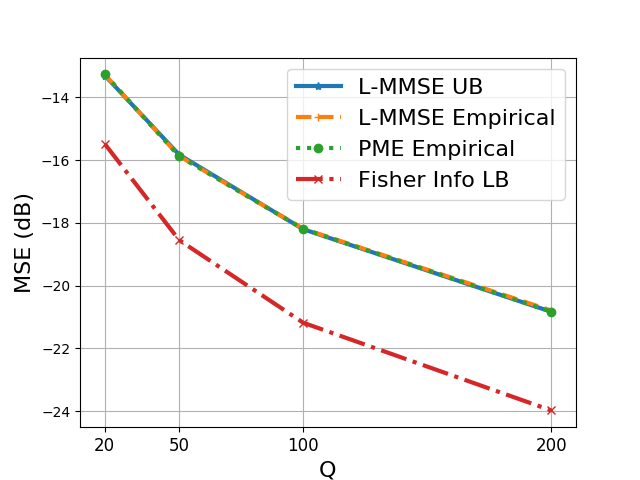}
} %\hspace{0.1cm}
\subfigure[$U=50$, SNR$=-10$dB.]{
\includegraphics[scale=0.28]{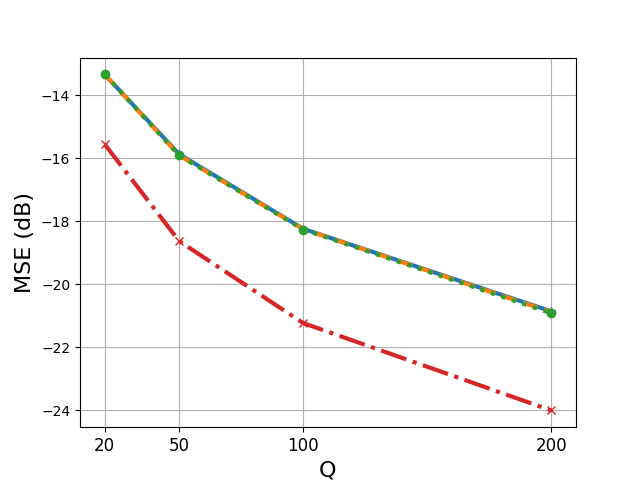}
} %\hspace{0.1cm}
\subfigure[$U=100$, SNR$=-10$dB.]{
\includegraphics[scale=0.28]{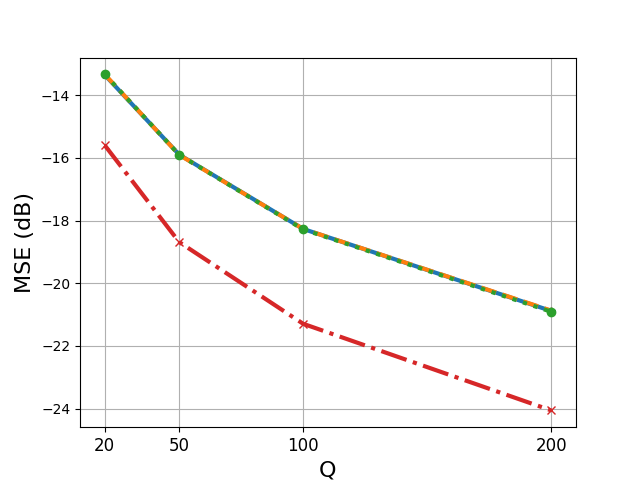}
}\\
\vspace{-0.4cm}
\subfigure[$U=20$, SNR$=0$dB.]{
\includegraphics[scale=0.28]{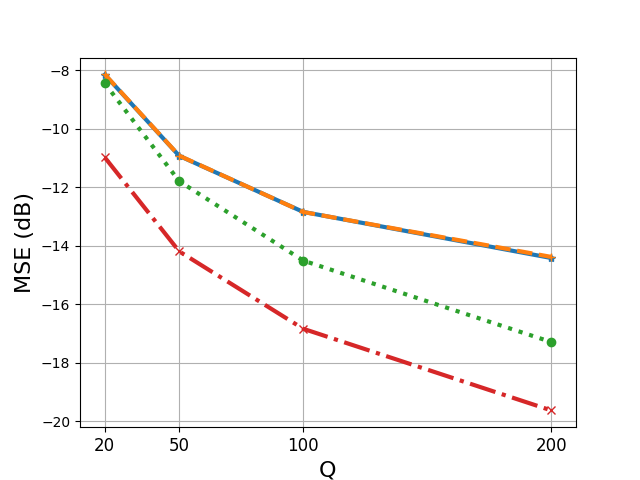}
} %\hspace{0.1cm}
\subfigure[$U=50$, SNR$=0$dB.]{
\includegraphics[scale=0.28]{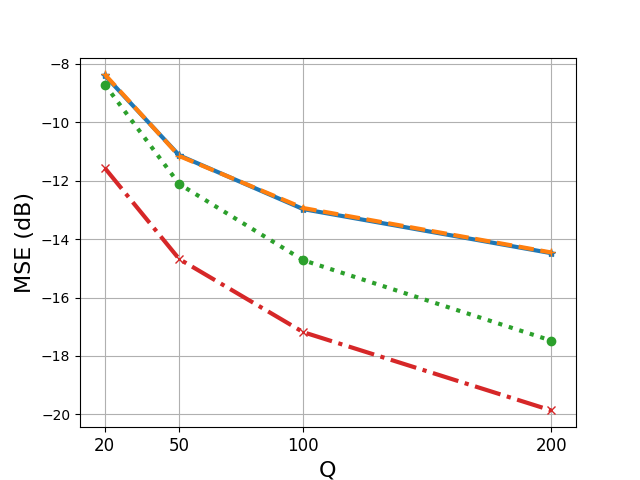}
} %\hspace{0.1cm}
\subfigure[$U=100$, SNR$=0$dB.]{
\includegraphics[scale=0.28]{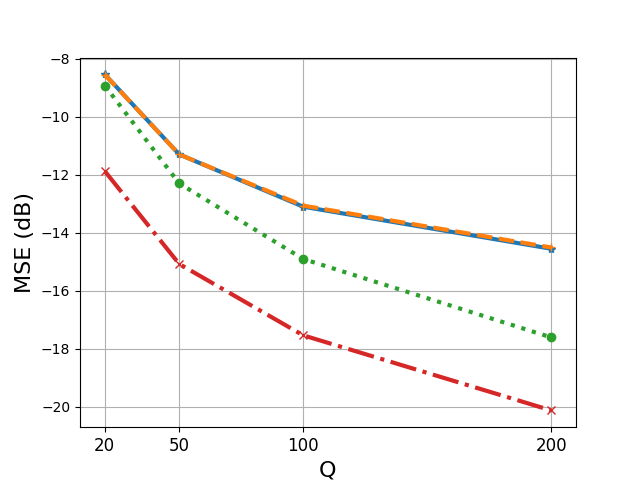}
}\\
\vspace{-0.4cm}
\subfigure[$U=20$, SNR$=10$dB.]{
\includegraphics[scale=0.28]{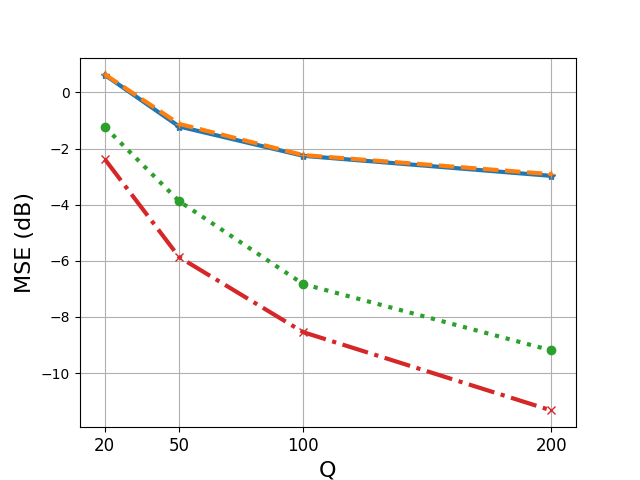}
} %\hspace{0.1cm}
\subfigure[$U=50$, SNR$=10$dB.]{
\includegraphics[scale=0.28]{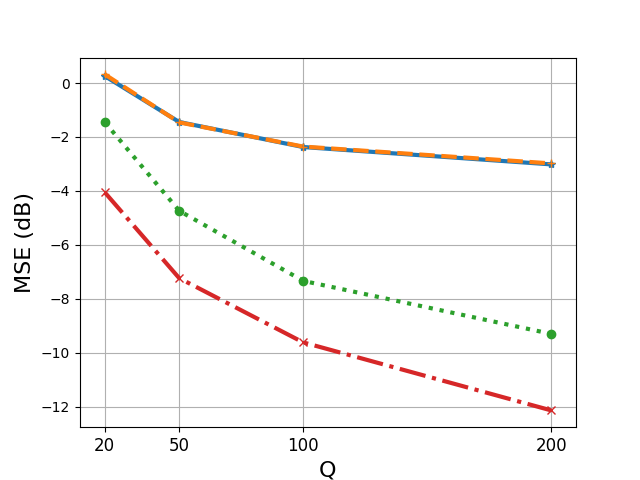}
} %\hspace{0.1cm}
\subfigure[$U=100$, SNR$=10$dB.]{
\includegraphics[scale=0.28]{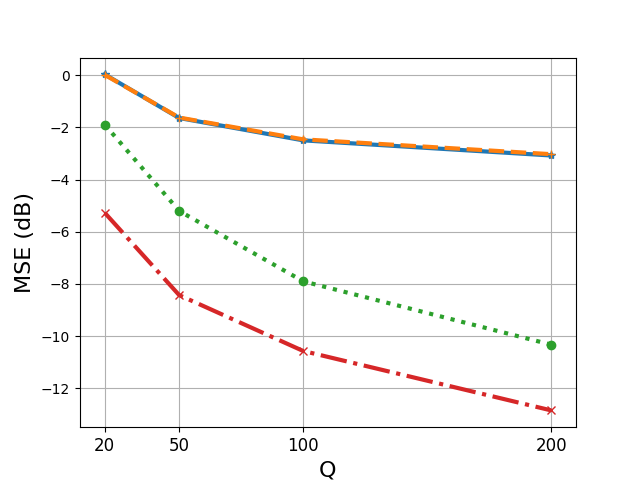}
}
\vspace{-0.2cm}
\caption{Empirical MSEs of the L-MMSE and PM estimators and the L-MMSE-based upper and Fisher information-based lower bounds on the MSE of the PM estimator for various SNR levels and problem sizes, when both user and item parameters are unknown.  We see that the upper bound is tight at low SNR and at all SNRs when the problem size is small.}
\vspace{-0.4cm}
\label{fig:compare}
\end{figure*}

\paragraph{Experimental Setup}
We vary the number of users $U \in \{ 20, 50, 100\}$ and the number of items $Q \in \{20, 50, 100, 200\}$. 
We generate the user ability and item difficulty parameters from zero-mean Gaussian distributions with variance $\sigma^2_x = \sigma^2_a = \sigma^2_d$.  We vary $\sigma^2_x$ so that the signal-to-noise ratio (SNR) corresponds to $\{ -10, 0, 10 \}$ decibels (dB).  
We then randomly generate the response from each user to each item, $Y_{u,i}$, according to \fref{eq:raschmodel}.  
We repeat these experiments for $1,000$ random instances of user and item parameters and responses, and report the averaged results.  

We compute the L-MMSE-based upper bound on the MSE of the PM estimator using \fref{thm:lmmseestimator} and the Fisher information-based lower bound using the method detailed in \cite{3plerr,scaleerr}.  Since the calculation of the Fisher information matrix requires the true values of the user ability and item difficulty parameters (which are to be estimated in practice), we use the PM estimates of these parameters instead.  
We also calculate the empirical parameter estimation MSEs of the L-MMSE and PM estimators.
To this end, we use a standard Gibbs sampling procedure \cite{albertchib}; we use the mean of the generated samples over $20,000$ iterations as the PM estimate after a burn-in phase of $10,000$ iterations.  We then use these estimates to calculate the empirical MSE.  

\begin{figure*}[t]
\vspace{-0.4cm}
\centering
\subfigure[SNR$=-10$dB.]{
\includegraphics[scale=0.28]{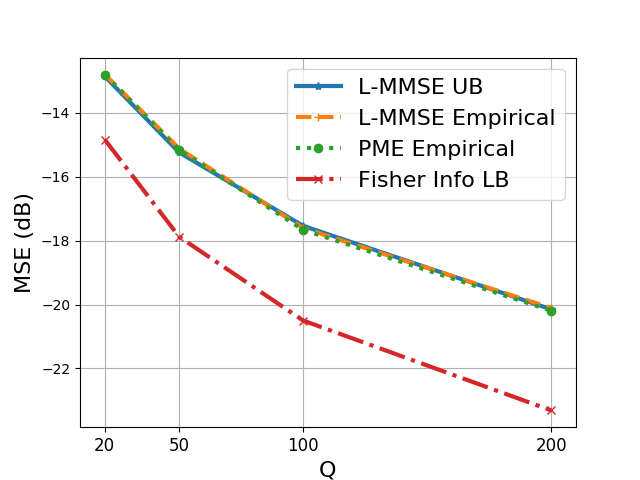}
} %\hspace{0.1cm}
\subfigure[SNR$=1$dB.]{
\includegraphics[scale=0.28]{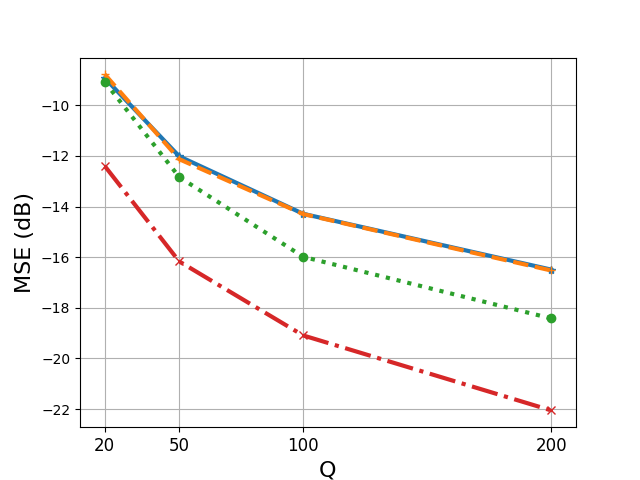}
} %\hspace{0.1cm}
\subfigure[SNR$=10$dB.]{
\includegraphics[scale=0.28]{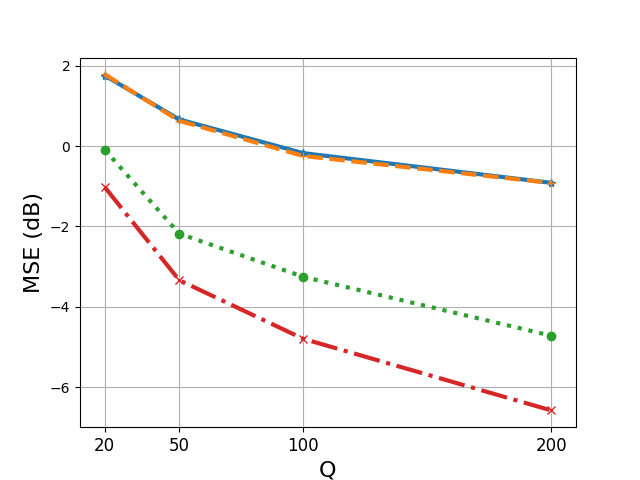}
}
\vspace{-0.2cm}
\caption{Empirical MSEs of the L-MMSE and PM estimators and the L-MMSE-based upper and Fisher Information-based lower bounds on the MSE of the PM estimator for various SNR levels and various problem sizes, when item parameters are known.  We see that the upper bound is tight at low SNR and at higher SNRs when the problem sizes are small.}
\vspace{-0.4cm}
\label{fig:knowmucompare}
\end{figure*}

\paragraph{Results and Discussion}
\fref{fig:compare} shows the empirical MSEs of the L-MMSE and PM estimators, together with the L-MMSE-based upper bound and the Fisher information-based lower bound on the MSE of the PM estimator, for every problem size and every SNR.  
First, we see that the analytical and empirical MSEs of the L-MMSE estimator match perfectly, which confirms that our analytical MSE expressions are \emph{exact}.  
We also see that for low SNR (i.e., the first row of \fref{fig:compare}), our L-MMSE upper bound on the MSE of the PM estimator is tight.  Moreover, at all noise levels, the L-MMSE-based upper bound is tighter at small problem sizes, while the Fisher information-based lower bound is tighter at very large problem sizes and at high SNR.  

These results confirm that our L-MMSE-based upper bound on the MSE is nonasymptotic, while the Fisher information-based lower bound is asymptotic and thus only tight at very large problem sizes.  Therefore, the L-MMSE-based upper bound is more practical than the Fisher information-based lower bound in real-world applications, especially for situations like the initial phase of CAT when the number of items a user has responded to is small.

\subsubsection{Case Two: Known item parameters}
\label{sec:knownexpt}

\paragraph{Experimental Setup}
In this experiment, we randomly generate the item parameters from the standard normal distribution ($\sigma^2_d = 1$) and treat these parameters as known; we then estimate the user ability parameters via \fref{thm:raschk}. 
The rest of the experimental setup remains unchanged.

\paragraph{Results and Discussion}
\fref{fig:knowmucompare} shows the empirical MSEs of the L-MMSE and PM estimators, together with the L-MMSE-based upper bound and the Fisher information-based lower bound on the MSE of the PM estimator, for every problem size and every SNR.  
We see that the analytical and empirical MSEs of the L-MMSE estimator match. 
We also see that the L-MMSE-based upper bound on the MSE is tighter than the Fisher information-based lower bound at low SNR levels ($-10$\,dB and $1$\,dB), and especially when the problem size is small (less than $50$ items).  
These results further confirm that our L-MMSE-based upper bound on the MSE is nonasymptotic, and is thus practical in the ``cold-start''  setting of recommender systems.

\begin{table*}[tp] 
\renewcommand{\arraystretch}{1.00}
\vspace{-0.4cm}
\centering
\caption{Mean and standard deviation of the prediction accuracy (ACC) for the L-MMSE, MAP, PM, and Logit-MAP estimators.}  \label{tbl:acc}
\vspace{0.1cm}
\scalebox{0.95}{ 
\begin{tabular}{@{}lcccc@{}}
\toprule
 & L-MMSE  & MAP & PM & Logit-MAP \\
\midrule
MT & $0.795 \pm 0.016 $ & $\bf 0.796 \pm 0.015 $ & $ 0.796 \pm 0.016 $ & $0.794 \pm 0.015 $ \\
SS & $\bf 0.860 \pm 0.007 $ & $0.859 \pm 0.007 $ & $0.859 \pm 0.007 $ & $0.859 \pm 0.010 $ \\
edX & $0.932 \pm 0.001 $ & $0.934 \pm 0.002 $ & $\bf 0.935 \pm 0.002 $ & $0.934 \pm 0.002 $ \\
ML & $\bf 0.715 \pm 0.004 $ & $0.713 \pm 0.004 $ & $0.713 \pm 0.004 $ & $0.714 \pm 0.004 $ \\
\bottomrule
\end{tabular}
}
\vspace{-0.4cm}
\end{table*}

\begin{table*}[tp] 
\renewcommand{\arraystretch}{1.00}
\centering
\caption{Area under the receiver operating characteristic curve (AUC) of the L-MMSE, MAP, PM, and Logit-MAP estimators.} \label{tbl:auc}
\vspace{0.1cm}
\scalebox{0.95}{ 
\begin{tabular}{@{}lcccc@{}}
\toprule
 & L-MMSE  & MAP & PM & Logit-MAP \\
\midrule
MT & $0.840 \pm 0.016 $  & $\bf 0.843 \pm 0.015 $ & $\bf 0.843 \pm 0.015 $ & $0.842 \pm 0.015 $ \\
SS & $0.800 \pm 0.014 $  & $\bf 0.803 \pm 0.013 $ & $\bf 0.803\pm 0.013 $ & $0.802 \pm 0.013 $ \\
edX & $0.900 \pm 0.004 $ & $\bf 0.909 \pm 0.004 $ & $\bf 0.909 \pm 0.004 $ & $\bf 0.909 \pm 0.004 $ \\
ML & $0.755 \pm 0.005 $ & $\bf 0.756 \pm 0.004 $ & $\bf 0.756 \pm 0.004 $ & $\bf 0.756 \pm 0.004 $ \\
\bottomrule
\end{tabular}
}
\vspace{-0.4cm}
\end{table*}

\subsection{Experiments with Real-World Data}
\label{sec:rwexpt}

% ================================================================================
% ================================================================================
% ================================================================================

We now test the performance of the proposed L-MMSE estimator using a variety of real-world datasets.  Since the noise model in real-world datasets is generally unknown, we also consider the  performance of MAP estimation using the inverse logit link function (Logit-MAP).
%, following \cite{linprobitciss}.
\vspace{-0.2cm}
\paragraph{Datasets}
We perform our experiments using a range of collaborative filtering datasets. 
These datasets are matrices that contain binary-valued ratings (or graded responses) of users (or students) to movies (or items).  For these datasets, we use the probit Rasch model.
%\footnote{For the details of the Rasch model and additional analytical results leveraging the specific structure of the design matrix $\bD$ under this model, please refer to the supplementary material.} 
%
The datasets include (i) ``MT'', which consists of students' binary-valued (correct/incorrect) graded responses to questions in a high-school algebra test, with $U = 99$ students' $3,366$ responses to $Q = 34$ questions, (ii) ``SS'', which consists of student responses in a signals and systems course, with $U = 92$ students' $5,692$ responses to $Q = 203$ questions, (iii) ``edX'', which consists of student responses in an edX course, with $U = 3241$ students' $177,181$ responses to $Q = 191$ questions, and (iv) ``ML'', a processed version of the \textit{ml-100k} dataset from the Movielens project \cite{movielens}, with $37,175$ integer-valued ratings by $U = 943$ users to $Q = 1152$ movies. 
We adopt the procedure used in \cite{1bitmc} to transform the dataset into binary values by comparing each rating to the overall average rating.
%Since the original \textit{ml-100k} dataset contains integer-valued ratings, we follow the procedure detailed in \cite{1bitmc} and transform it into binary-valued ratings by comparing each rating to the overall average rating across the entire dataset. %; we also trim the dataset to remove users and movies with only few ratings.  
%
\vspace{-0.2cm}
\paragraph{Experimental Setup}
We evaluate the prediction performance of the L-MMSE, MAP, PM, and Logit-MAP estimators using ten-fold cross validation.  
We randomly divide the entire dataset into ten equally-partitioned folds (of user-item response pairs), leave out one fold as the held-out testing set and use the other folds as the training set.  We then use the training set to estimate the learner abilities $a_u$ and item difficulties $d_i$, and use these estimates to predict user responses on the test set.  We tune the prior variance parameter $\sigma_x^2$ using a separate validation set (one fold in the training set).  %We repeat each experiment for $20$ random partitions of each dataset.  
%In all experiments, we fix $\sigma_w^2 = 1$.
%
To assess the performance of these estimators, we use two common metrics in binary classification problems: prediction accuracy (ACC), which is simply the portion of correct predictions, and area under the receiver operating characteristic curve (AUC) \cite{accauc}.  Both metrics have range in $[0,1]$, with larger values indicating better predictive performance.  
%and the covariance matrix of the offset parameters as $\bC_\bmx = \sigma_x^2 \bI $.

\vspace{-0.2cm}
\paragraph{Results and Discussion}

Tables~\ref{tbl:acc} and~\ref{tbl:auc} show the mean and standard deviation of the performance of each estimator on both metrics across each fold.
We observe that the performance of the considered estimators are comparable on the ACC metric, while the L-MMSE estimator performs slightly worse than the MAP, PM, and Logit-MAP estimators for most datasets on the AUC metric.  
%As the size of the dataset grows larger, e.g., for the MT and SS datasets, the performance gap between MAP/PM and L-MMSE gets larger. 
%
%; however, for the edX and ML datasets that contain a large number of observations, the performance of LS is on par with other estimators.
%Moreover, 

We find it quite surprising that a well-designed linear estimator performs on par with more sophisticated nonlinear estimators on these real-world datasets.  We also note that the L-MMSE estimator is more computationally efficient than the PM estimator.  As an example, on the MT and ML datasets, one run of the L-MMSE estimator takes $0.23$s and $79$s, respectively, while one run of the PM estimator takes $1.9$s and $528$s ($2,000$ and $10,000$ iterations required for convergence) on a standard laptop computer.  These observations suggest that the L-MMSE estimator is computationally efficient and thus scales favorably to large datasets. 
%the L-MMSE estimator can be implemented efficiently and scales well to larger datasets, which is in stark contrast to the PM estimator. 
%
%Furthermore, since the design matrix $\bD$ is given for each of these experiments, our analytical results provide insights into the MSE of the vector $\bmx$ recovered by our linear estimators (see Theorem~\ref{thm:rasch}). %\cs{as shown in the rasch analysis section?}

% Recall that our experiments on synthetic data show that at the performance of all estimators are similar at very low SNR;  Therefore, since their performance is virtually indistinguishable on real-world datasets, it is likely that if we assume real-world datasets follow the probit likelihood model, then they have very low SNRs.  

%%%%%%%%%%%%%%%%%%%%%%%%%%%%%%%%%%%%%%%%%%%%%%%%%%%%%%%%%%

\vspace{-0.2cm}
\section{Conclusions}

We have generalized a recently proposed linear estimator for probit regression and applied the method to the classic Rasch model in item response analysis.  We have shown that the L-MMSE estimator enables an exact, closed-form, and nonasymptotic MSE analysis, which is in stark contrast to existing analytical results which are asymptotic, probabilistic, or loose.  As a result, we have shown that the nonasymptotic, L-MMSE-based upper bound on the parameter estimation error of the PM estimator under the Rasch model can be tighter than the common Fisher information-based asymptotic lower bound, especially in practical settings.  An avenue of future work is to apply our analysis to models that are more sophisticated than the Rasch model, e.g., the latent factor model in \cite{sparfa}. 
%

% ================================================================================
% ================================================================================
% ================================================================================

% !TEX root = 18ICML_linrasch.tex
%%%%%%%%%%%%%%%%%%%%%%%%%%%%%%

\appendix

\section{Proof of \fref{thm:raschk}}
\label{app:pfknown}

Let $\bmz = \bD \bmx + \bmm + \bmw$.  Thus, $\bmz \distas \mathcal{N}(\bD \bar{\bmx} + \bmm, \bD \bC_\bmx \bD^T + \bI) := \mathcal{N}(\bar{\bmz},\bC_\bmz)$.  
The L-MMSE estimator for $\bmx$ has the general form of $\hat{\bmx}^\text{L-MMSE} = \bW \bmy + \bmb$, where $\bW = \bE \bC_{\bmy}^{-1}$ and $\bmb = \bar{\bmx} - \bW \bar{\bmy}$, with 
\begin{align*}
\bC_{\bmy} \! = \! \Ex{}{(\bmy \! - \! \bar{\bmy})(\bmy \! - \! \bar{\bmy})^T} \! = \! \Ex{}{\bmy \bmy^T} \! - \! \bar{\bmy} \bar{\bmy}^T \! := \! \widetilde{\bC}_\bmy \! - \! \bar{\bmy} \bar{\bmy}^T
\end{align*}
and 
\begin{align*}
\bE = \Ex{}{(\bmy - \bar{\bmy})(\bmx - \bar{\bmx})^T} \! = \! \Ex{}{\bmy \bmx^T} \! - \! \bar{\bmy} \bar{\bmx}^T \! := \! \widetilde{\bE} \! - \!  \bar{\bmy} \bar{\bmx}^T.
\end{align*}
We need to evaluate three quantities, $\bar{\bmy}$, $\widetilde{\bC}_\bmy$, and $\widetilde{\bE}$.  

We start with $\bar{\bmy}$.  Its $i$th entry is given by  
\begin{align*}
\bar{y}_i & = \int_{-\infty}^\infty \sign(z_i) \mathcal{N}(z_i; \bar{z}_i, [\bC_\bmz]_{i,i}) \mathrm{d}z_i \\
& = \!-\!\!\int_{-\infty}^0 \!\!\! \mathcal{N}(z_i; \bar{z}_i, [\bC_\bmz]_{i,i}) \mathrm{d}z_i \!+\!\! \int_0^\infty \!\!\! \mathcal{N}(z_i; \bar{z}_i, [\bC_\bmz]_{i,i}) \mathrm{d}z_i \\
& = \Phi \left(\frac{\bar{z}_i}{\sqrt{[\bC_\bmz]_{i,i}}} \right) - \Phi\left(-\frac{\bar{z}_i}{\sqrt{[\bC_\bmz]_{i,i}}}\right).
\end{align*}
Next, we calculate $\widetilde{\bC}_\bmy$.  Its $(i,j)$th entry is given by $[\widetilde{\bC}_\bmy]_{i,j} = $
\begin{align*}
& \int_{-\infty}^\infty \int_{-\infty}^\infty \sign(z_i) \sign(z_j) \mathcal{N} \Big( \Big[ \begin{array}{c} z_i \\ z_j \end{array} \Big]; \\
& \quad \quad  \Big[ \begin{array}{c} \bar{z}_i \\ \bar{z}_j \end{array} \Big], \Big[ \begin{array}{cc} [\bC_\bmz]_{i,i} & [\bC_\bmz]_{i,j} \\ {[\bC_\bmz]_{j,i}} & [\bC_\bmz]_{j,j} \end{array} \Big] \Big) \mathrm{d}z_j \mathrm{d}z_i \\
& \overset{(a)}{=} \int_{-\infty}^\infty \int_{-\infty}^\infty \sign\left(\frac{z_i + \bar{z}_i}{\sqrt{[\bC_\bmz]_{i,i}}}\right) \sign\left(\frac{z_j + \bar{z}_j}{\sqrt{[\bC_\bmz]_{j,j}}}\right)  \\
& \quad \quad \mathcal{N} \Big( \Big[ \begin{array}{c} z_i \\ z_j \end{array} \Big]; \bZero, \Big[ \begin{array}{cc} 1 & \rho \\ \rho & 1 \end{array} \Big] \Big) \mathrm{d}z_j \mathrm{d}z_i \\
& = \! \underbrace{\int_{-\infty}^{-\frac{\bar{z}_i}{\sqrt{[\bC_\bmz]_{i,i}}}} \!\! \int_{-\infty}^{-\frac{\bar{z}_j}{\sqrt{[\bC_\bmz]_{j,j}}}} \!\! \mathcal{N} \Big( \Big[ \begin{array}{c} z_i \\ z_j \end{array} \Big]; \bZero, \Big[ \begin{array}{cc} 1 & \rho \\ \rho & 1 \end{array} \Big] \Big) \mathrm{d}z_j \mathrm{d}z_i}_{v_1} \\
& \quad + \! \underbrace{\int_{-\frac{\bar{z}_i}{\sqrt{[\bC_\bmz]_{i,i}}}}^\infty \!\! \int_{-\frac{\bar{z}_j}{\sqrt{[\bC_\bmz]_{j,j}}}}^\infty \!\! \mathcal{N} \Big( \Big[ \begin{array}{c} z_i \\ z_j \end{array} \Big]; \bZero, \Big[ \begin{array}{cc} 1 & \rho \\ \rho & 1 \end{array} \Big] \Big) \mathrm{d}z_j \mathrm{d}z_i}_{v_2} \\
& \quad - \! \underbrace{\int_{-\infty}^{-\frac{\bar{z}_i}{\sqrt{[\bC_\bmz]_{i,i}}}} \!\!\! \int_{-\frac{\bar{z}_j}{\sqrt{[\bC_\bmz]_{j,j}}}}^\infty \!\!\! \mathcal{N} \Big( \Big[ \begin{array}{c} z_i \\ z_j \end{array} \Big]; \bZero, \Big[ \begin{array}{cc} 1 & \rho \\ \rho & 1 \end{array} \Big] \Big) \mathrm{d}z_j \mathrm{d}z_i}_{v_3} \\
& \quad - \! \underbrace{\int_{-\frac{\bar{z}_i}{\sqrt{[\bC_\bmz]_{i,i}}}}^\infty \!\! \int_{-\infty}^{-\frac{\bar{z}_j}{\sqrt{[\bC_\bmz]_{j,j}}}} \!\!\! \mathcal{N} \! \Big( \Big[ \begin{array}{c} z_i \\ z_j \end{array} \Big]; \bZero, \Big[ \begin{array}{cc} 1 & \rho \\ \rho & 1 \end{array} \Big] \Big) \mathrm{d}z_j \mathrm{d}z_i}_{v_4} \\
& \overset{\text{(b)}}{=} 2 (v_1 + v_2) - 1\\
& = 2\Bigg(\Phi_2\left(\frac{\bar{z}_i}{\sqrt{[\bC_\bmz]_{i,i}}}, \frac{\bar{z}_j}{\sqrt{[\bC_\bmz]_{j,j}}}, \rho\right) \\ 
& \quad + \Phi_2\left(-\frac{\bar{z}_i}{\sqrt{[\bC_\bmz]_{i,i}}}, -\frac{\bar{z}_j}{\sqrt{[\bC_\bmz]_{j,j}}}, \rho\right)\Bigg) - 1,
\end{align*}
where we have used (a) change of variable $\frac{z_i - \bar{z}_i}{\sqrt{[\bC_\bmz]_{i,i}}} \rightarrow z_i$ and (b) the fact that $v_1 + v_2 + v_3 + v_4 = 1$.  
The computation of $\widetilde{\bE}$ follows from that in \cite{linprobitciss} and is omitted.

\section{Proof of \fref{thm:rasch}}
\label{app:pfunknown}

%\begin{proof}[Proof of Theorem~5]
Recall that the expression for the MSE is $\tr(\bC_\bmx - \bE^T \bC_\bmy^{-1} \bE)$, the critical part is to evaluate $\bE^T \bC_\bmy^{-1} \bE $.  We begin by evaluating $\bC_\bmy^{-1}$. 
For the Rasch model, we have $\bD=[\bOne_{Q}\kron\bI_{U\times U},\bI_{Q\times Q}\kron\bOne_U]$. Therefore, since $\bC_\bmx = \sigma^2_x \bI_{U+Q}$, we have
\begin{align*}
\bC_\bmz & = \bD \bC_\bmx \bD^T + \bI_{UQ \times UQ} = \sigma^2_x \bD \bD^T + \bI_{UQ \times UQ} \\
& = \sigma^2_x [\bOne_{Q}\kron\bI_{U\times U} \quad \bI_{Q\times Q}\kron\bOne_U]  \left[ \begin{array}{c} \bOne^T_{Q}\kron\bI_{U\times U} \\ \bI_{Q\times Q}\kron\bOne_U^T \end{array} \right] \\
& \quad + \bI_{UQ \times UQ} \\
& = \sigma^2_x (\bOne_{Q \times Q} \kron \bI_{U\times U} + \bI_{Q\times Q} \kron \bOne_{U \times U}) + \bI_{UQ \times UQ},
\end{align*}
where $\bOne_{U \times U}$ denotes an all-one matrix with size $U \times U$. 
Therefore, we see that the $UQ \times UQ$ matrix $\bC_\bmz$ consists of three parts: (i) $Q$ copies of the all-ones matrix $\sigma^2_x \bOne_{U \times U}$ in its diagonal $U \times U$ blocks, (ii) copies of the matrix $\sigma^2_x \bI_{U\times U}$ in every other off-diagonal $U \times U$ block, plus (iii) a diagonal matrix $\bI_{UQ \times UQ}$.  Therefore, its diagonal elements are $2\sigma^2_x + 1$ and its non-zero off-diagonal elements are $\sigma^2_x$.  

As detailed in \citep[(7)]{linprobitciss}, one can show that 
\begin{align*}
\bC_{\bmy} & =  \frac{2}{\pi} \arcsin(\mathrm{diag}( \mathrm{diag}(\bC_\bmz)^{-1/2}) \, \bC_\bmz  \\
& \quad \times \mathrm{diag}(\mathrm{diag}(\bC_\bmz)^{-1/2})),
\end{align*}
we have that the term inside the $\arcsin$ function has the same structure as $\bC_\bmz$, with diagonal entries of $1$ and non-zero off-diagonal entries as $\frac{\sigma^2_x}{2\sigma^2_x + 1}$.  Therefore, $\bC_{\bmy}$ also has the same structure, with diagonal entries of $1$ and non-zero off-diagonal entries as 
\begin{align*}
s= \frac{2}{\pi} \arcsin\!\left(\frac{\sigma^2_x}{2\sigma^2_x + 1}\right).
\end{align*}

Since $\bC_\bmy^{-1}$ satisfies $\bC_\bmy \bC_\bmy^{-1} = \bI_{UQ \times UQ}$, it is easy to see that the entries of $\bC_\bmy^{-1}$ only contain three distinct values (denoted by $a$, $b$, and $c$), and consists of two parts: (i) $Q$ copies of a $U \times U$ matrix with $a$ on its diagonal, $b$ everywhere else, in its diagonal blocks, and (ii) copies of a $U \times U$ matrix with $c$ on its diagonal, $d$ everywhere else, in its other blocks.  We next compute $a$, $b$, $c$, and $d$.

The first column of $\bC_\bmy^{-1}$ is given by 
\begin{align*}
[a, b \bOne_{1 \times U-1}, c, d\bOne_{1 \times U-1}, c, d\bOne_{1 \times U-1}, \ldots]^T.
\end{align*}
Since its inner product with the first row of $\bC_\bmy$ is one (since $\bC_\bmy \bC_\bmy^{-1} = \bI_{UQ \times UQ}$), we get
\begin{align*}
a + (U-1)sb + (Q-1)sc  = 1. 
\end{align*}
Similarly, its inner products with the second, $(U+1)-th$, and $(U+2)$-th rows are all zero; this gives 
\begin{align*}
 sa + ((U-2)s+1)b + (Q-1)sd & = 0, \\
 sa + ((Q-2)s+1)c + (U-1)sd & = 0, \\
 sb + sc + ((U+Q-4)s+1)d & = 0.
\end{align*}
Solving the linear system given by these four equations results in
\begin{align} \label{eq:abcd}
\notag a & \!= \! \frac{(3U^2 \!\! + \! 3Q^2 \!\!-\! U^2Q \!\!-\! UQ^2 \!\!+\! 8UQ \!-\! 15U \!-\! 15 Q \!+\! 20) s^3}{r} \\
\notag & \quad + \frac{(- U^2 - Q^2 - 3UQ + 11 U + 11 Q - 22) s^2}{r}  \\
\notag & \quad + \frac{(- 2 U -2 Q + 8) s - 1}{r} \\
\notag b & \!=\! \frac{(UQ + Q^2 - 3U - 5Q + 8)s^3 \!+\! (U + 2Q - 6)s^2 \!+\! s}{r} \\
\notag c & \!=\! \frac{(UQ + U^2 - 5U - 3Q + 8)s^3 \!+\! (2U + Q - 6)s^2 \!+\! s}{r} \\
d & \!=\! \frac{-(U+Q - 4)s^3 - 2s^2}{r},
\end{align}
where 
\begin{align*}
r \!=\! (2s\!-\!1)((U\!-\!2)s\!+\!1)((Q\!-\!2)s\!+\!1)((Q\!+\!U\!-\!2)s\!+\!1).
\end{align*} 
Now, let $\bA$ be the $N \times N$ matrix with $c$ on its diagonal and $d$ everywhere else, $\bB$ denote the matrix with $a-c$ on its diagonal and $b-d$ everywhere else, we can write $\bC_\bmy^{-1}$ as 
\begin{align} \label{eq:cykron}
\bC_\bmy^{-1} = \bOne_{Q \times Q} \kron \bA + \bI_{Q \times Q} \kron \bB.
\end{align}

\sloppy
Our second task is to evaluate $\bE$.  Since
\begin{align*}
\bE & =  \sqrt{\frac{2}{\pi}} \mathrm{diag}(\mathrm{diag}(\bC_\bmz)^{-1/2})\bD\bC_\bmx 
 = \sqrt{\frac{2}{\pi}} \frac{\sigma^2_x}{\sqrt{2 \sigma^2_x + 1}} \bD \\
& = \frac{\sigma^2_x}{\sqrt{2 \sigma^2_x + 1}} [\bOne_{Q}\kron\bI_{U\times U} \;\; \bI_{Q\times Q}\kron\bOne_U], 
\end{align*}
we have
\begin{align*}
& \bE^T \bC_\bmy^{-1} \bE =  \frac{2}{\pi} \frac{\sigma^4_x}{2 \sigma^2_x + 1} \left[ \begin{array}{c} \bOne^T_{Q}\kron\bI_{U\times U} \\ \bI_{Q\times Q}\kron\bOne_U^T \end{array} \right] \\
& \quad \times (\bOne_{Q \times Q} \kron \bA + \bI_{Q \times Q} \kron \bB) [\bOne_{Q}\kron\bI_{U\times U} \;\; \bI_{Q\times Q}\kron\bOne_U] \\ 
& = \frac{2}{\pi} \frac{\sigma^4_x}{2 \sigma^2_x + 1} Q ( Q\bA + \bB),
\end{align*}
where we have used $(\bX \kron \bY) (\bU \kron \bV) = (\bX \bU) \kron (\bY \bV)$.  

Therefore, the value of entry $(1,1)$ in $\bE^T \bC_\bmy^{-1} \bE$, i.e., the MSE of the user ability parameter estimates, is given by
\begin{align*} 
& \frac{2}{\pi} \frac{\sigma^4_x}{2 \sigma^2_x + 1} Q (a + (Q-1)c) = \\
&  \sigma_x^2 \! \left(\!1 \!-\! \frac{2}{\pi} \! \frac{\sigma^2_x}{2 \sigma^2_x + 1} \! \frac{sQ(Q+U-3)+1}{(s(Q-2)+1)(s(Q+U-2)+1)}\!\right)\!, 
\end{align*}
where we have used \fref{eq:abcd}, thus completing the proof.

\section*{Acknowledgments}
C.~Studer was supported in part by Xilinx, Inc.\ and by the US National Science Foundation (NSF) under grants ECCS-1408006, CCF-1535897,  CCF-1652065, and CNS-1717559.
%
%\section*{Acknowledgments}
%%
%The authors would like to thank S. Jacobsson and G. Durisi for crucial discussions and insights on Bussgang's decomposition. 

% ================================================================================
% ================================================================================
% ================================================================================

\balance
%\vspace{-0.15cm}
\bibliographystyle{icml2018}

{\small
\begin{spacing}{1.0} % dirty hack!!! :-)
\bibliography{confs-jrnls,publishers,lmmse}
\end{spacing}
}

\balance

\end{document}